\documentclass[11pt,twoside]{article}
\usepackage[preprint]{acl}
\usepackage{iftex}
\ifPDFTeX
  \usepackage{times}
  \usepackage[T1]{fontenc}
  \usepackage[utf8]{inputenc}
\else
  \usepackage{fontspec}
\fi
\usepackage{microtype}
\usepackage{amsmath}
\usepackage{amssymb}
\usepackage{booktabs}
\usepackage{graphicx}

\title{How Well Do Pre-Generation Answer Error Probes Transfer?\\Controlled Evidence in Gemma-3}

\author{
  Ali Jafar \quad
  Abdul Ghani \quad
  Muhammad Tayyab \quad
  Maryam Bashir \\
  FAST School of Computing \\
  National University of Computer and Emerging Sciences (FAST-NUCES), Lahore, Pakistan \\
  {\normalfont\small \texttt{ali.jafar.fast@gmail.com} (A.\ Jafar) \quad \texttt{maryam.bashir@nu.edu.pk} (M.\ Bashir)}
}

\begin{document}
\maketitle

\begin{abstract}
Can a linear probe on a language model's prompt-time hidden state predict which upcoming answers will be wrong on an unseen corpus? We evaluate frozen Gemma-3-4B-IT representations immediately before decoding across nine medical QA corpora ($5{,}848$ binary items; $4{,}973$ cross-corpus) and five MRQA extractive QA corpora ($3{,}235$ items; $2{,}750$ cross-corpus). Single-source probes lose error-ranking discrimination under shift ($0.624 \rightarrow 0.558$ AUROC medical; $0.759 \rightarrow 0.709$ MRQA). With multi-source leave-one-corpus-out training, retaining $250$ dense features raises AUROC over all $2{,}560$ coordinates from $0.627$ to $0.689$ on medical QA (cross-source stability $F$) and from $0.763$ to $0.807$ on MRQA (pooled $F$), exceeding sparse autoencoder latents and lexical baselines. Dimension-matched random masks ($0.677$ medical, $0.788$ MRQA), directional-sign checks, and zero-penalty optimizer controls reveal that capacity reduction and optimization regularization drive most gains, while risk extrapolation adds no AUROC benefit over matched $\beta = 0$. Disjoint covariance alignment helps medical pooled selection ($+0.035$ AUROC, $-0.013$ AP) but hurts MRQA, and cross-task transfer averages near chance ($0.500$, $0.489$).
\end{abstract}

\section{Introduction}

Detecting when a large language model (LLM) will produce an incorrect answer before generation begins offers a practical trigger for retrieval or human review without generating multiple rollouts \citep{kadavath2022mostly,ji2024risk,moreno2025noanswer}. Suppose a classifier monitors a clinical assistant or reading-comprehension pipeline trained on existing question logs, and is then deployed on a new hospital portal or QA collection. For that monitor to remain reliable, its error ranking must survive shifts in corpus composition, domain terminology, and prompt style.

Recent studies establish that prompt-time hidden states encode signals predictive of upcoming answer accuracy on trivia and knowledge benchmarks \citep{ji2024risk,moreno2025noanswer}, whereas linear truth or verification probes often degrade across corpora, medical registers, and task families \citep{mishra2026medical,duan2026transfer}. Yet prior evaluations typically test probes either on candidate answers already placed in the prompt \citep{azaria2023internal,mishra2026medical} or compare sparse and adapted probes without matched dimension, random-subset, surface-text, or optimizer controls \citep{gulrajani2021domainbed,kantamneni2025sae}. Our study addresses that gap in a fixed-backbone empirical audit on frozen \texttt{google/gemma-3-4b-it}: probes observe only the prompt representation immediately prior to the first answer token and predict whether Gemma's own greedy continuation will later fail reference grading.

We organize the investigation around three cumulative research questions, where each stage tests a concrete design choice against its tightest matched control:
\begin{enumerate}
    \item \textbf{Single-source transfer:} How sharply does error-ranking performance degrade when a classifier trained on one corpus is deployed on another within medical QA or extractive QA?
    \item \textbf{Multi-source coordinate selection and controls:} On an unseen corpus, does pooling sources and filtering dense coordinates (or moving to sparse autoencoder latents) beat full-state probing—and do those gains survive random-250 subspaces, signed cross-source consistency checks, surface-text models, and a zero-penalty optimizer control for variance risk extrapolation (V-REx)?
    \item \textbf{Target access and cross-task boundaries:} When a small unlabeled batch or a handful of labeled target examples become available, does closed-form covariance alignment (CORAL) or few-shot reweighting help cleanly, and does error ranking survive crossing between open-ended medical QA and passage-grounded extractive QA?
\end{enumerate}

Our controlled results qualify several optimistic expectations. First, shrinking the dense hidden space from $2{,}560$ to $250$ features improves held-out error ranking cleanly over full-space probing across both benchmarks, with stability-ranked $F$-scores leading on medical AUROC ($0.6891$, $+0.0619$) and pooled $F$-scores leading on MRQA ($0.8068$, $+0.0440$). Second, matched controls overturn simple attributions: ten random 250-feature masks already reach $0.6775$ (medical) and $0.7883$ (MRQA), unsigned stability features reverse sign frequently across medical sources ($10.7\%$ unanimous direction vs.\ $42.7\%$ for signed mean-$t$, which reaches $0.6879$ AUROC and top medical average precision $0.4405$), and V-REx's variance penalty underperforms the exact same clipped-Adam optimizer at $\beta=0$ in \emph{both} benchmarks ($0.6688$ vs.\ $0.6905$ medical; $0.7901$ vs.\ $0.7953$ MRQA). Third, unlabeled alignment on a disjoint adaptation batch gives mixed AUROC/AP trade-offs on medical QA ($0.6565 \rightarrow 0.6910$ for Pooled $F$, AP $0.440 \rightarrow 0.427$) and hurts MRQA, while zero-shot cross-task transfer averages near chance.

\section{Experimental Setup, Corpora, and Audits}

To isolate representation and adaptation decisions cleanly, we fix the base model, extraction layer, prompt token position, and primary linear solver family throughout all controlled comparisons. Across every benchmark, answers are produced via deterministic greedy decoding from frozen unquantized BF16 \texttt{google/gemma-3-4b-it} (revision \texttt{093f9f38}). During the single prompt-prefill forward pass, we extract the $2{,}560$-dimensional output of transformer block 17 (zero-indexed language-model layer) at the final formatted prompt token—before any generated answer token exists—and project the same state through the open Gemma Scope~2 residual sparse autoencoder (\texttt{layer\_17\_width\_\allowbreak 16k\_l0\_medium}, $16{,}384$ nonnegative latents) \citep{gemma2025report,mcdougall2025scope}. Prompts, decoding caps, and model details appear in Appendix~\ref{sec:appendix_d}.

\subsection{Benchmarks, Error Labels, and Provenance Audit}

\paragraph{Medical QA (9 Corpora).}
We evaluate nine English medical QA datasets spanning NIH health pages, drug inquiries, patient forums, African clinical Q\&A, USMLE exams, and telemedicine dialogues: MedQuAD \citep{benabacha2019medquad}, MedicationQA \citep{benabacha2019medication}, MedRedQA \citep{nguyen2023medredqa}, AskD-English \citep{askd}, AfriMed-QA short-answer \citep{nimo2025afrimed}, OpenMedQA \citep{openmedqa}, MedQA-open \citep{nachane2024medqaopen}, and HealthCareMagic and iCliniq \citep{li2023chatdoctor}. We draw $650$ queries per corpus ($5{,}850$ total) partitioned into train, validation, disjoint reserved adaptation, and test splits. Frozen NF4-quantized \texttt{Qwen/Qwen2.5-7B-Instruct} \citep{yang2025qwen} compares each generated answer against dataset references using a structured clinical rubric (Appendix~\ref{sec:appendix_d}), labeling errors as $y=1$, correct answers as $y=0$, and excluding $2$ \texttt{uncertain} verdicts ($1$ MedicationQA test row; $1$ MedQA-open train row).

Across all $5{,}848$ resulting binary items, $2{,}921$ train ($833$ errors), $882$ validation ($234$ errors), and $1{,}170$ test rows ($319$ errors, $27.3\%$) form the audited SHA256-verified leave-one-corpus-out archive ($4{,}973$ rows), while $875$ binary items remain untouched in the disjoint reserved adaptation pool ($97$--$99$ per corpus; MedQuAD $322/98/99/131$ splits reflect source group balancing). As shown in Table~\ref{tab:corpora}, Gemma's error prevalence varies nearly twelvefold across medical portals, from $6.2\%$ ($8/130$ test errors) on MedRedQA and iCliniq to $71.5\%$ ($93/130$) on OpenMedQA.

Because several medical collections derive from related web or exam repositories, we audited exact and fuzzy string overlap across all $36$ corpus pairs using whitespace-stripped normalization and character $3$--$5$-gram / word $1$--$2$-gram TF-IDF cosine similarity (Appendix~\ref{sec:appendix_a}). Exact normalization identifies $1$ overlapping question between MedRedQA (reserved adaptation split) and AskD-English (train split)—which single-source pairwise transfer excludes symmetrically from both ($96$ reserve and $324$ train), whereas multi-source LODO keeps AskD-English at $325$ train rows because the MedRedQA duplicate sits outside LODO splits in the reserved adaptation pool. More importantly, character $n$-gram scanning exposes \textbf{$29$ near-duplicate pairs} between OpenMedQA and MedQA-open at cosine $\ge 0.85$ ($8$ at $\ge 0.95$, maximum $0.9840$), including paraphrased USMLE vignettes crossing between OpenMedQA training and MedQA-open test splits. All other $34$ corpus pairs have zero pairs above $0.75$ char cosine. To account for this shared MedQA and AskDocs lineage explicitly, our medical suite includes a \emph{paired family holdout} removing both related corpora simultaneously.

\begin{table}[t]
\centering
\scriptsize
\setlength{\tabcolsep}{3pt}
\resizebox{\columnwidth}{!}{%
\begin{tabular}{lrrrrr}
\toprule
Corpus & Train & Val & Res. & Test & Test Err.\ (\%) \\
\midrule
MedQuAD & 322 & 98 & 99 & 131 & 42 (32.1\%) \\
MedicationQA & 325 & 98 & 97 & 129 & 30 (23.3\%) \\
MedRedQA & 325 & 98 & 97* & 130 & 8 (6.2\%) \\
AfriMed-QA & 325 & 98 & 97 & 130 & 35 (26.9\%) \\
OpenMedQA & 325 & 98 & 97 & 130 & 93 (71.5\%) \\
MedQA-open & 324 & 98 & 97 & 130 & 74 (56.9\%) \\
HealthCareMagic & 325 & 98 & 97 & 130 & 19 (14.6\%) \\
AskD-English & 325* & 98 & 97 & 130 & 10 (7.7\%) \\
iCliniq & 325 & 98 & 97 & 130 & 8 (6.2\%) \\
\midrule
\textbf{MedQA Binary (9)} & \textbf{2,921} & \textbf{882} & \textbf{875} & \textbf{1,170} & \textbf{319 (27.3\%)} \\
\midrule
SQuAD & 322 & 98 & 97 & 130 & 11 (8.5\%) \\
HotpotQA & 322 & 98 & 97 & 130 & 24 (18.5\%) \\
TriviaQA & 322 & 98 & 97 & 130 & 27 (20.8\%) \\
NaturalQuestions & 322 & 98 & 97 & 130 & 46 (35.4\%) \\
SearchQA & 322 & 98 & 97 & 130 & 72 (55.4\%) \\
\midrule
\textbf{MRQA Binary (5)} & \textbf{1,610} & \textbf{490} & \textbf{485} & \textbf{650} & \textbf{180 (27.7\%)} \\
\bottomrule
\end{tabular}%
}
\caption{Sample counts and held-out test error prevalence ($y=1$) across all 9 medical QA corpora ($5{,}848$ binary items; $4{,}973$ LODO train/val/test) and 5 MRQA extractive corpora ($3{,}235$ items; $2{,}750$ LODO train/val/test). *Pairwise single-source deduplication drops 1 shared question across MedRedQA reserve ($96$) and AskD-English train ($324$).}
\label{tab:corpora}
\end{table}

\paragraph{Extractive Reading Comprehension (MRQA 2019, 5 Corpora).}
To test whether findings generalize beyond open-ended medical recall to passage-grounded reading comprehension, we sample five MRQA~2019 development corpora \citep{fisch2019mrqa}: SQuAD, TriviaQA-web, SearchQA, HotpotQA, and NaturalQuestionsShort ($647$ items per corpus, $3{,}235$ total; $322$ train, $98$ validation, $97$ reserved, $130$ test). Partitioning respects normalized passage groups, yielding zero cross-split passage or question leakage and zero cross-corpus question overlap across all five datasets. Gemma receives the passage and question under one standardized template and generates up to $48$ greedy tokens. Following MRQA answer normalization, an answer is coded as an error ($y=1$) when not an exact match and maximum token F1 falls below $0.50$ ($180/650$ test errors, $8.5\%$--$55.4\%$), and we evaluate grading sensitivity across F1 $<0.25$, F1 $<0.75$, and exact mismatch rules.

\subsection{Evaluation Metrics and Statistical Scope}
Following \citet{calderon2024robustness}, let $SS$, $ST$, and $TT$ denote AUROC of a source probe on its own test set, on an unseen target $T$, and of an in-domain target-trained reference on $T$, defining \emph{Source-Drop} ($SD = SS - ST$) and \emph{Target-Drop} ($TD = TT - ST$). Because error prevalence spans $6.2\%$ to $71.5\%$, we report threshold-free average precision (AP; \citealp{davis2006pr}) beside AUROC in every primary comparison; AP baseline precision equals class prevalence whereas random AUROC stays $0.50$. AUROC measures error \emph{ranking}, not probability calibration or deferral utility.

All macro averages weight each target corpus equally. Paired $95\%$ percentile bootstrap intervals resample held-out test rows $B=2{,}000$ times within each corpus stratified by class (seed $42$ plus target index) and average paired score differences across targets \citep{efron1979bootstrap}. Intervals quantify test-sample uncertainty conditional on these fixed corpora and fitted probes rather than variability across future corpus collections.

\subsection{Matched Training, Feature Filters, and Controls}

\paragraph{Single-Source vs.\ Multi-Source LODO Protocols.}
In single-source evaluation, balanced $L_2$ logistic regression (LIBLINEAR solver, $C \in \{0.01, 0.1, 1, 10\}$) selects $C$ by 3-fold stratified grouped cross-validation inside source training rows (using passage groups inside MRQA with in-fold scaling) and refits on source train+validation. In multi-source leave-one-domain-out (LODO) evaluation, one corpus $T$ is held out and all remaining $M$ corpora ($M=8$ medical; $M=4$ MRQA) form pooled source training and validation sets. Candidate regularization strengths $C \in \{0.01, 0.1, 1.0, 10.0\}$ (and linear SVM / Ridge baselines) are selected strictly on the \emph{unweighted macro mean of the $M$ within-source validation AUROCs} \citep{gulrajani2021domainbed}, ties broken toward stronger regularization, and then refit on combined source train+validation rows. Target-trained references tune $C$ using only $T$'s own train and validation splits. Dense vectors are standardized to zero mean and unit variance using source-fit statistics; sparse SAE features scale to unit variance without centering to preserve sparsity.

\paragraph{Coordinate Selection and Dimension Controls.}
Fixing the feature budget at $K=250$ (selected after an initial medical grid sweep over $\{50, 100, 250\}$), we compare three source-train ANOVA $F$-statistic rankers \citep{fisher1925} against two signed source-$t$ rankers and three dimension-matched controls:
\begin{itemize}
    \item \textbf{Pooled $F$}: one-way ANOVA $F(x_{\text{train}, :, j}, y_{\text{train}})$ computed on pooled source training rows.
    \item \textbf{Mean $F$ and Stable $F$}: let $F_e(j)$ be the within-source ANOVA $F$ on source $e$. \emph{Mean $F$} ranks by $\frac{1}{M}\sum_{e=1}^M F_e(j)$, whereas \emph{Stable $F$} divides by cross-source variability, $\bar{F}(j) / (\operatorname{std}_e(F_e(j)) + 10^{-5})$. Note that $F_e$ discards sign polarity.
    \item \textbf{Signed Mean-$t$ and Signed Stable-$F$}: to test whether directional sign reversal invalidates unsigned stability, we compute the two-sample Welch $t$-statistic $t_e(j)$ (errors minus correct answers) within each source and rank by $|\bar{t}(j)|$ (\emph{Signed Mean-$t$}) or $|\bar{t}(j)| / (\operatorname{std}_e(t_e(j)) + 10^{-5})$ (\emph{Signed Stable-$F$}).
    \item \textbf{Random 250, PCA 250, and Text Controls}: ten random 250-coordinate masks (seeds 42--51, averaged across fitted probes), $250$ source-fitted PCA components \citep{pearson1901}, character/word prompt length, and TF-IDF word $1$--$2$-grams (up to $5{,}000$ terms; \citealp{salton1988}), evaluated both at fixed $C=1.0$ and validation-tuned $C$.
\end{itemize}

\paragraph{V-REx Optimizer Control, Reserved CORAL, and Few-Shot Budget.}
For V-REx \citep{krueger2021rex}, a linear head optimizes the unweighted mean of balanced per-source BCE losses ($\frac{1}{M}\sum_{e=1}^M \mathcal{L}_e$) plus penalty $\beta \operatorname{Var}_e(\mathcal{L}_e)$ using Adam ($\text{lr}=10^{-3}$), gradient clipping ($1.0$), and early stopping on validation macro AUROC: tuning $C \in \{0.01..10\}$ and $\beta \in \{0, 0.1, 1, 10, 50\}$ over $150$ epochs (checkpointed every $15$, biased variance) on medical QA, and fixed weight decay $10^{-4}$ with $\beta \in \{0, 1, 10\}$ over $100$ epochs (unbiased variance) on MRQA. Crucially, the $\beta=0$ control matches every setting identically within each benchmark to isolate optimizer effects from the variance penalty. For closed-form CORAL \citep{sun2016coral}, source standardized features in the selected $250$-d subspace are transformed by $A = C_S^{-1/2} C_T^{1/2}$ ($\gamma=1.0$) using \emph{only the disjoint unlabeled reserved adaptation split} ($N=97$--$99$ medical, $97$ MRQA) while keeping identical coordinates and tuned $C$ fixed against the unaligned probe. Exploratory adaptation variants (transductive test-covariance CORAL, few-shot target adaptation $K \in \{0..50\}$, DEAP Brier calibration, Mixture-of-Probes, LRCP, DPO ranking, and Kernel MMD) are audited separately in Appendix~\ref{sec:appendix_c} so protocol access differences remain transparent.

\section{Single-Source Transfer Degrades Across Corpora}

\begin{table}[t]
\centering
\scriptsize
\setlength{\tabcolsep}{3.5pt}
\resizebox{\columnwidth}{!}{%
\begin{tabular}{llccccc}
\toprule
Benchmark (Shifts) & Repr. & $SS$ & $ST$ & $TT$ & $TD$ & $TD > 0$ \\
\midrule
MedQA (72) & Dense & 0.6241 & 0.5579 & 0.6241 & $+0.0662$ & 49/72 \\
MedQA (72) & SAE & 0.6110 & 0.5646 & 0.6110 & $+0.0464$ & 41/72 \\
MRQA (20, grouped) & Dense & 0.7594 & 0.7090 & 0.7594 & $+0.0505$ & 15/20 \\
\bottomrule
\end{tabular}%
}
\caption{Single-source directed transfer across 72 medical pairs and 20 passage-grouped MRQA pairs. Positive $TD = TT - ST$ indicates an in-domain probe outperforms cross-corpus transfer. (Earlier ungrouped MRQA SAE CV yielded $0.7477 \rightarrow 0.6992$, $15/20$ $TD>0$.)}
\label{tab:pairwise}
\end{table}

To answer \textbf{RQ1}, Table~\ref{tab:pairwise} shows that error-ranking performance consistently falls when a probe trained on a single corpus is transferred elsewhere within the same QA setting. In medical QA, dense probes average $0.6241$ AUROC in-domain—reaching $0.8457$ on AfriMed-QA and $0.8373$ on MedQuAD, yet dropping near chance on informal forums such as MedRedQA ($0.4191$)—and lose $+0.0662$ AUROC on average across $72$ directed shifts ($0.5579$, ranging from $0.3779$ for OpenMedQA$\rightarrow$HealthCareMagic to $0.8084$ for MedQuAD$\rightarrow$AfriMed-QA), with in-domain probes beating transferred probes on $49/72$ cells. SAE latents behave similarly ($0.6110 \rightarrow 0.5646$, $41/72$ positive drops).

In passage-based MRQA, within-corpus error ranking is higher overall ($0.7594$ dense), yet cross-corpus transfer across $20$ passage-grouped pairs still drops by $+0.0505$ AUROC ($0.7090$, range $0.5259$ SQuAD$\rightarrow$HotpotQA to $0.8341$ TriviaQA$\rightarrow$SearchQA, $15/20$ positive drops). Because single-corpus training rarely generalizes reliably, the natural next decision is to pool multiple sources and examine whether multi-source training—with or without sparse dictionaries or feature selection—resolves the drop.

\section{Multi-Source Selection and What Controls Reveal}

\begin{table*}[t]
\centering
\small
\setlength{\tabcolsep}{4.5pt}
\resizebox{\textwidth}{!}{%
\begin{tabular}{lcccccl}
\toprule
Method (Feature Budget / Solver) & Med.\ AUC & Med.\ AP & MRQA AUC & MRQA AP & Target Labels & Protocol Role \\
\midrule
All dense features ($2{,}560$; LR) & 0.6272 & 0.3749 & 0.7628 & 0.5521 & 0 & Primary full-state baseline \\
Pooled SAE latents ($16{,}384$; LR) & 0.6352 & 0.4143 & 0.7207 & 0.4914 & 0 & Matched sparse representation \\
Target-trained Dense (LR ref.) & 0.6365 & 0.4124 & 0.7594 & 0.5424 & Train+Val & In-domain supervised reference \\
\midrule
Random $250$ mean (10 seeds; LR) & 0.6775 & 0.4208 & 0.7883 & 0.5987 & 0 & Dimension-matched random floor \\
PCA $250$ components (LR) & 0.6413 & 0.3865 & 0.7778 & 0.5520 & 0 & Unsupervised projection control \\
Pooled $F$ ($250$; LR) & 0.6565 & 0.4399 & \textbf{0.8068} & \textbf{0.6420} & 0 & Global class separation ranker \\
Mean $F$ ($250$; LR) & 0.6711 & 0.4190 & 0.7986 & 0.6179 & 0 & Unweighted source mean $F$ \\
Stable $F$ ($250$; LR) & 0.6891 & 0.4233 & 0.7803 & 0.5749 & 0 & Unsigned cross-source stability \\
Signed Stable-$F$ ($250$; LR) & 0.6771 & 0.4080 & 0.7912 & 0.6078 & 0 & Signed cross-source $t$ stability \\
Signed Mean-$t$ ($250$; LR) & 0.6879 & \textbf{0.4405} & 0.7978 & 0.6229 & 0 & Signed mean effect ($t$) \\
\midrule
V-REx (Adam + clip, tuned $\beta\ge 0$) & 0.6688 & 0.4293 & 0.7901 & 0.6225 & 0 & Risk variance penalty \\
V-REx Control ($\beta=0$, same opt.) & \textbf{0.6905} & 0.4298 & 0.7953 & 0.6150 & 0 & Matched zero-penalty optimizer \\
\midrule
TF-IDF $1$--$2$gram ($C=1$ / val-tuned) & 0.5834 / 0.5818 & 0.3301 / 0.3376 & 0.5429 / 0.5469 & 0.3189 / 0.3357 & 0 & Lexical surface control \\
Prompt Length ($C=1$ / val-tuned) & 0.5542 / 0.5201 & 0.3431 / 0.3168 & 0.5238 / 0.5164 & 0.3309 / 0.3323 & 0 & Surface length control \\
\bottomrule
\end{tabular}%
}
\caption{Matched multi-source LODO macro AUROC and Average Precision (AP) across 9 medical corpora ($1{,}170$ test rows; $319$ errors) and 5 MRQA corpora ($650$ test rows; $180$ errors). Bold values mark the single highest score per column (with Stable-$F$ leading fixed coordinate selectors at $0.6891$ Med.\ AUC). Additional matched full-space baselines: MedQA Target SAE ($0.6402$ AUC / $0.4156$ AP), Dense SVM/Ridge ($0.5957$ / $0.5947$), SAE SVM/Ridge ($0.6022$ / $0.6050$); MRQA Target SAE ($0.7477$ / $0.5706$).}
\label{tab:main_lodo}
\end{table*}

\begin{table*}[t]
\centering
\small
\setlength{\tabcolsep}{4pt}
\resizebox{\textwidth}{!}{%
\begin{tabular}{llcccccc}
\toprule
Method & Comparator & Med.\ $\Delta$AUC [95\% CI] & Med.\ $\Delta$AP [95\% CI] & MRQA $\Delta$AUC [95\% CI] & MRQA $\Delta$AP [95\% CI] \\
\midrule
Random 250 mean & All features & \textbf{+0.050 [+0.020, +0.081]} & \textbf{+0.046 [+0.019, +0.071]} & \textbf{+0.026 [+0.000, +0.053]} & \textbf{+0.047 [+0.000, +0.087]} \\
Pooled $F$ & All features & $+0.029$ $[-0.011, +0.070]$ & \textbf{+0.065 [+0.022, +0.105]} & \textbf{+0.044 [+0.008, +0.078]} & \textbf{+0.090 [+0.029, +0.137]} \\
Stable $F$ & All features & \textbf{+0.062 [+0.021, +0.103]} & \textbf{+0.048 [+0.013, +0.087]} & $+0.018$ $[-0.017, +0.050]$ & $+0.023$ $[-0.036, +0.075]$ \\
Signed Mean-$t$ & All features & \textbf{+0.061 [+0.024, +0.100]} & \textbf{+0.066 [+0.027, +0.098]} & \textbf{+0.035 [+0.001, +0.069]} & \textbf{+0.071 [+0.009, +0.119]} \\
\midrule
Stable $F$ & Random 250 mean & $+0.012$ $[-0.018, +0.040]$ & $+0.002$ $[-0.021, +0.031]$ & $-0.008$ $[-0.035, +0.015]$ & $-0.024$ $[-0.062, +0.013]$ \\
Pooled $F$ & Random 250 mean & $-0.021$ $[-0.048, +0.006]$ & $+0.019$ $[-0.007, +0.048]$ & $+0.018$ $[-0.007, +0.040]$ & \textbf{+0.043 [+0.005, +0.072]} \\
Signed Mean-$t$ & Random 250 mean & $+0.010$ $[-0.015, +0.036]$ & $+0.020$ $[-0.004, +0.040]$ & $+0.009$ $[-0.016, +0.031]$ & $+0.024$ $[-0.015, +0.054]$ \\
Stable $F$ & Pooled $F$ & $+0.033$ $[-0.006, +0.069]$ & $-0.017$ $[-0.055, +0.024]$ & $\mathbf{-0.026\ [-0.051,\ -0.001]}$ & $\mathbf{-0.067\ [-0.110,\ -0.016]}$ \\
\midrule
V-REx ($\beta=0$) & All features & \textbf{+0.063 [+0.024, +0.105]} & $+0.055$ (point) & $+0.033$ $[-0.003, +0.067]$ & $+0.063$ (point) \\
V-REx (tuned $\beta\ge 0$) & V-REx ($\beta=0$) & $-0.022$ $[-0.048, +0.004]$ & $-0.001$ (point) & $-0.005$ (point) & $+0.007$ (point) \\
Best Filter & Tuned TF-IDF & \textbf{+0.107 [+0.042, +0.173]} & \textbf{+0.086 [+0.026, +0.142]} & \textbf{+0.260 [+0.196, +0.325]} & \textbf{+0.306 [+0.221, +0.374]} \\
\bottomrule
\end{tabular}%
}
\caption{Direct paired bootstrap differences ($B=2{,}000$ stratified test draws conditional on fitted models). Bold entries mark intervals strictly excluding zero (MRQA Random 250 vs.\ All bounds round from $+0.0005$ AUROC and $+0.0003$ AP). Best Filter vs.\ Tuned TF-IDF evaluates Stable-$F$ on Medical QA and Pooled-$F$ on MRQA.}
\label{tab:contrasts}
\end{table*}

\begin{figure*}[t]
    \centering
    \includegraphics[width=\textwidth]{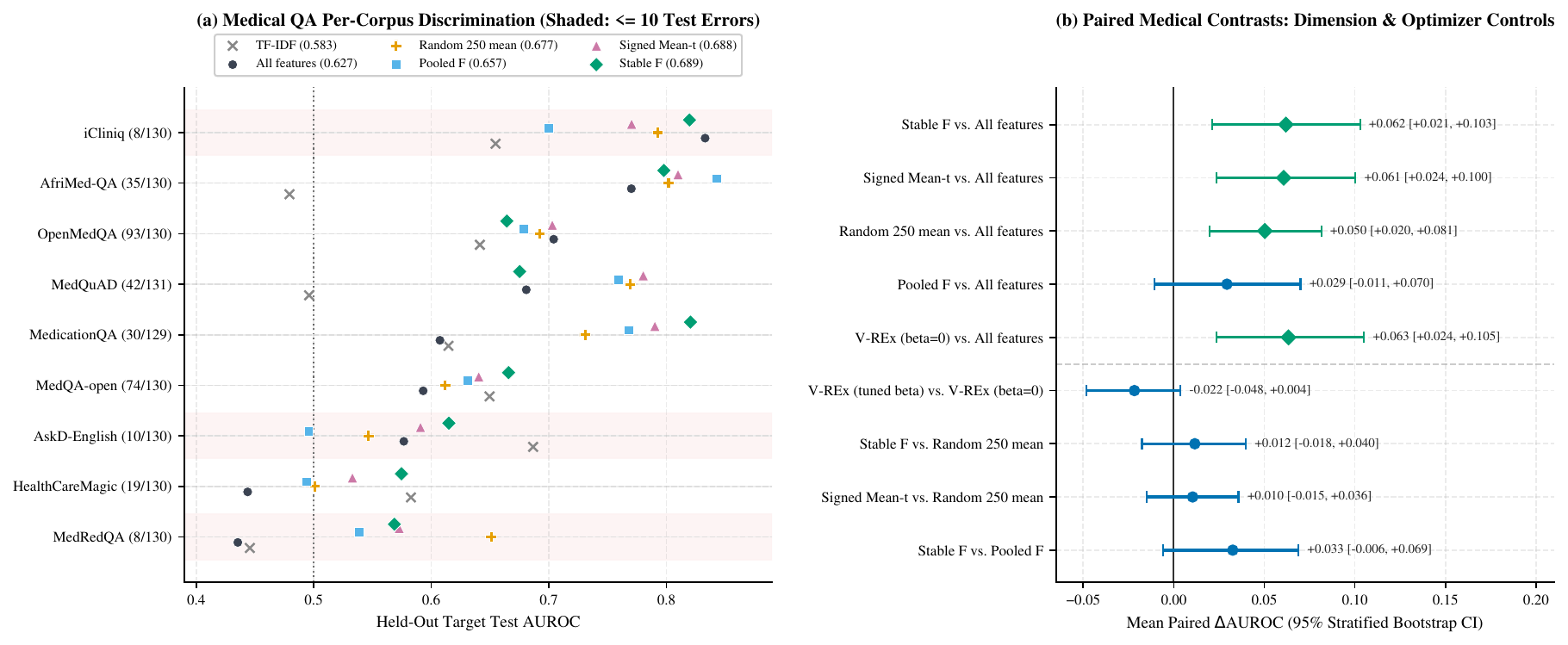}
    \caption{Medical QA LODO evaluation across 9 held-out corpora. \textbf{(a)} Per-corpus test AUROC (targets sorted by full-probe AUROC, showing held-out error counts; shaded rows mark portals with $\le 10$ test errors). \textbf{(b)} Paired $\Delta$AUROC forest plot: shrinking from $2{,}560$ to $250$ features (including random masks) and gradient-clipped Adam ($\beta=0$) clear the full baseline, whereas Stable $F$ vs.\ Random 250 and V-REx vs.\ $\beta=0$ do not.}
    \label{fig:medqa_lodo}
\end{figure*}

\subsection{Naive Pooling, Sparse Autoencoders, and Feature Reduction}

Before adopting complex adaptation, we ask whether simply pooling multiple training corpora or expanding into sparse monosemantic latents solves cross-corpus generalization (Table~\ref{tab:main_lodo}). In medical QA, pooling eight sources raises macro AUROC from single-source $0.5579$ to $0.6272$ ($0.3749$ AP), nearly matching supervised target training ($0.6365$; difference $+0.0093$, CI $[-0.054, +0.077]$) and outperforming linear SVM ($0.5957$) and Ridge ($0.5947$). On low-support targets such as iCliniq ($8/130$ test errors), pooling sharply beats fragile in-domain target tuning ($0.8330$ vs.\ $0.4785$), whereas on MRQA pooling ($0.7628$) ties target training ($0.7594$). Replacing $2{,}560$ dense activations with $16{,}384$ Gemma Scope~2 SAE latents under identical tuning yields $0.6352$ medical AUROC ($+0.0081$, CI $[-0.043, +0.062]$) and $0.7207$ MRQA AUROC ($-0.0421$, CI $[-0.090, +0.010]$). Sparse dictionary expansion thus brings no reliable benefit over dense activations in either benchmark \citep{kantamneni2025sae}.

Retaining only $250$ dense coordinates changes both benchmarks (Tables~\ref{tab:main_lodo}--\ref{tab:contrasts}, Figures~\ref{fig:medqa_lodo} and~\ref{fig:mrqa_lodo}; full per-corpus AUROC and AP appear in Appendix~\ref{sec:appendix_b}, Tables~\ref{tab:per_fold_auc}--\ref{tab:medqa_ap}). In medical QA, \emph{Stable $F$} raises AUROC from $0.6272$ to $0.6891$ ($+0.0619$, CI $[+0.0213, +0.1028]$, winning $6/9$ corpora with $+0.2131$ on MedicationQA) and raises AP from $0.3749$ to $0.4233$ ($+0.0484$, CI $[+0.0128, +0.0873]$). On MRQA, by contrast, \emph{Pooled $F$} leads cleanly at $0.8068$ AUROC ($+0.0440$, CI $[+0.0083, +0.0783]$) and $0.6420$ AP ($+0.0899$, CI $[+0.0289, +0.1365]$), beating \emph{Stable $F$} ($0.7803$ AUROC, $0.5749$ AP) in paired comparisons on both metrics ($-0.0265$ AUROC, CI $[-0.0513, -0.0007]$; $-0.0671$ AP, CI $[-0.1099, -0.0160]$).

\subsection{Random Subsets, Direction Reversals, and Family Holdouts Challenge Naive Claims}

Does the medical improvement come from the cross-source stability formula specifically, or simply from fitting probes in a $250$-dimensional space instead of $2{,}560$? Direct dimension-matched controls isolate the mechanism (Table~\ref{tab:contrasts}):
\begin{itemize}
    \item \textbf{Random 250 features rival ANOVA on AUROC:} Averaged over ten random $250$-coordinate masks, \emph{Random 250} improves significantly over all $2{,}560$ features on both medical QA ($0.6775$ AUROC, $+0.0503$, CI $[+0.0199, +0.0814]$; $0.4208$ AP) and MRQA ($0.7883$, $+0.0255$, CI $[+0.0005, +0.0529]$). Against this dimension-matched floor, medical \emph{Stable $F$} leads by only $+0.0116$ AUROC ($[-0.0176, +0.0398]$) and $+0.0025$ AP. On MRQA, \emph{Pooled $F$}'s AUROC edge over \emph{Random 250} ($+0.0184$, $[-0.0072, +0.0405]$) also crosses zero, while its \textbf{AP gain clears zero cleanly ($+0.0433$, $[+0.0046, +0.0721]$)}.
    \item \textbf{Unsigned stability does not enforce consistent error polarity:} Because one-way $F_e$ squares group differences, a feature can score highly under \emph{Stable $F$} while separating errors in opposite directions across sources. Checking source-wise Welch $t$-statistics directly (Appendix Table~\ref{tab:sign_diag}) reveals that only $10.7\%$ of medical \emph{Stable $F$} coordinates keep a unanimous sign across all 8 source corpora (mean $|\overline{\operatorname{sign}}(t_e)| = 0.465$, lower than $0.599$ for \emph{Pooled $F$}). Replacing unsigned $F$ with \emph{Signed Stable-$F$} increases unanimous directional agreement to $58.0\%$ ($0.887$ mean consensus) on medical QA and $100\%$ on MRQA, yet scores $0.6771$ on medical QA—indistinguishable from \emph{Random 250} ($-0.0004$, $[-0.0223, +0.0199]$). Dropping the denominator std penalty via \emph{Signed Mean-$t$} maintains $42.7\%$ unanimous ($0.811$ consensus) agreement while achieving top medical AP ($0.4405$, beating all features cleanly on both AUROC $+0.0607$ and AP $+0.0656$) and $0.7978$ AUROC ($0.6229$ AP) on MRQA.
    \item \textbf{Provenance holdout sensitivity:} When holding out both AskDocs corpora together (MedRedQA and AskD-English) or both MedQA-derived corpora together (OpenMedQA and MedQA-open), mean AUROCs across those four targets remain close to single-corpus holdouts ($0.5867$ full vs.\ $0.5876$ Pooled $F$, $0.6104$ Stable $F$, and $0.6248$ Mean $F$; Appendix Table~\ref{tab:family_holdout}), with all paired family-minus-single differences spanning zero cleanly.
\end{itemize}

\begin{figure*}[t]
    \centering
    \includegraphics[width=\textwidth]{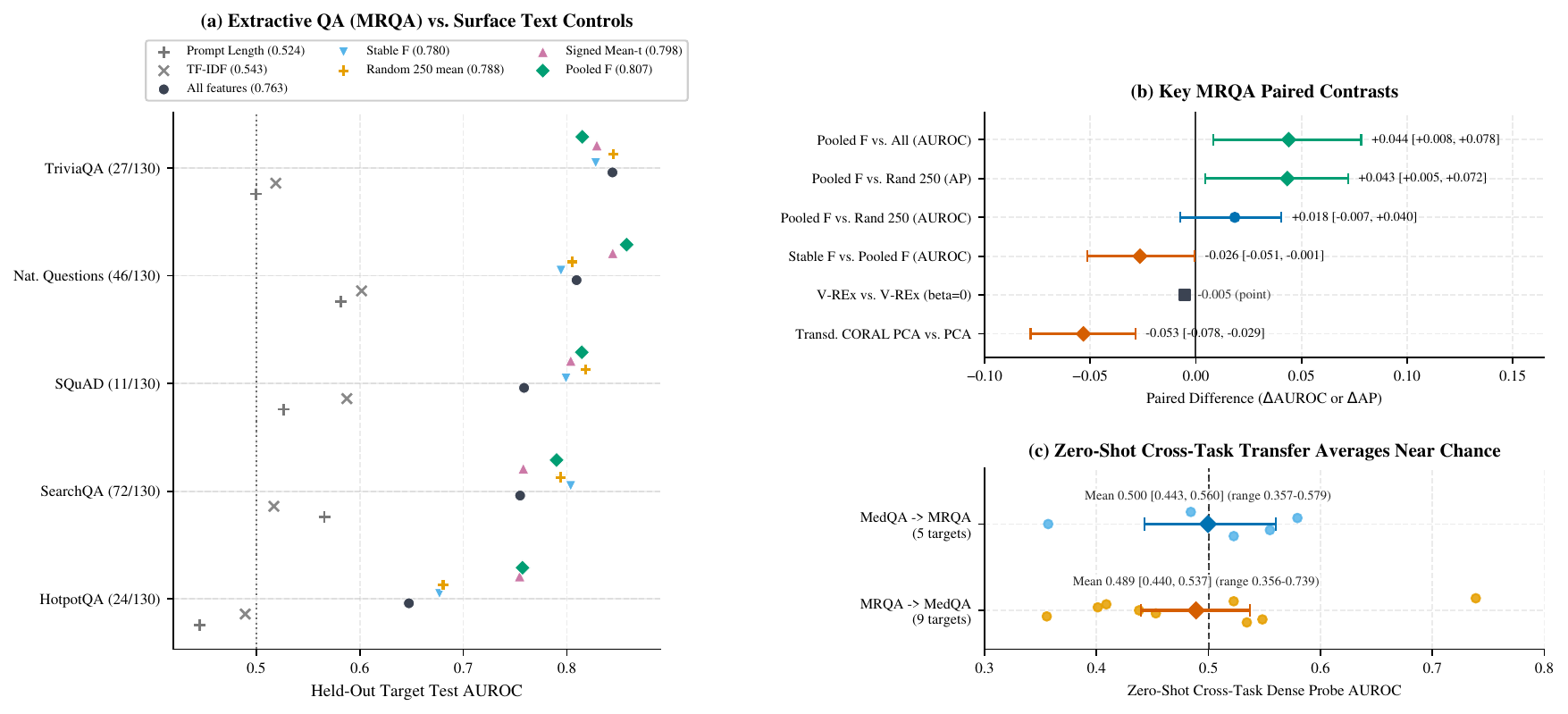}
    \caption{Extractive QA (MRQA) generalization and cross-task transfer boundary. \textbf{(a)} Pre-generation probes ($0.763$--$0.807$ AUROC) outperform prompt-length ($0.524$) and TF-IDF ($0.543$) surface controls on every corpus. \textbf{(b)} Paired MRQA contrasts confirm Pooled $F$ outperforms Full features (AUROC/AP), Stable $F$ (AUROC/AP), and Random 250 (AP), while tuned V-REx ($\beta\ge 0$) and transductive CORAL fail to improve. \textbf{(c)} Bidirectional zero-shot transfer between Medical QA and MRQA averages near chance with wide target scatter.}
    \label{fig:mrqa_lodo}
\end{figure*}

\subsection{Surface Text Controls and Why V-REx Controls Matter}

Surface prompt text predicts errors only weakly (Table~\ref{tab:main_lodo}). On MRQA, validation-tuned TF-IDF ($0.5469$) and prompt length ($0.5164$) sit near chance, while \emph{Pooled $F$} leads tuned TF-IDF by $+0.2599$ AUROC ($[+0.1958, +0.3249]$). On medical QA, TF-IDF reaches $0.5834$ fixed ($0.5818$ tuned)—exceeding unfiltered full dense probing on AskD-English ($0.6867$ vs.\ $0.5767$) and HealthCareMagic ($0.5827$ vs.\ $0.4438$) at fixed $C=1.0$ ($0.6367$ and $0.5657$ validation-tuned) so that unfiltered \emph{All features} minus TF-IDF ($+0.0453$, $[-0.0228, +0.1089]$) crosses zero, whereas \emph{Stable $F$} ($+0.1072$, $[+0.0421, +0.1733]$) and \emph{Signed Mean-$t$} ($+0.1060$ point delta) clear tuned TF-IDF by over $+0.10$ AUROC.

Finally, matched $\beta=0$ optimization controls resolve the apparent success of invariant risk extrapolation. Initial unclipped gradient descent for V-REx collapsed to $0.5000$ AUROC; adding gradient clipping, Adam, and early validation checkpointing lifts tuned V-REx ($\beta\ge 0$; selecting $\beta^*=0$ on $3/9$ medical folds) to $0.6688$ AUROC ($0.4293$ AP) on medical QA and $0.7901$ ($0.6225$ AP) on MRQA. Yet running the exact same optimization loop with fixed $\beta=0$ reaches \textbf{$0.6905$ AUROC ($0.4298$ AP)} on medical QA and \textbf{$0.7953$ AUROC ($0.6150$ AP)} on MRQA—outperforming tuned V-REx in both benchmarks ($\Delta = -0.0217$, $[-0.0484, +0.0037]$ on medical QA; $-0.0052$ on MRQA). Early-stopped gradient-clipped optimization, rather than cross-environment risk variance regularization, explains the improvement over LIBLINEAR pooled regression \citep{gulrajani2021domainbed}.

\section{Target Access and Cross-Task Boundaries}

\begin{figure*}[t]
    \centering
    \includegraphics[width=\textwidth]{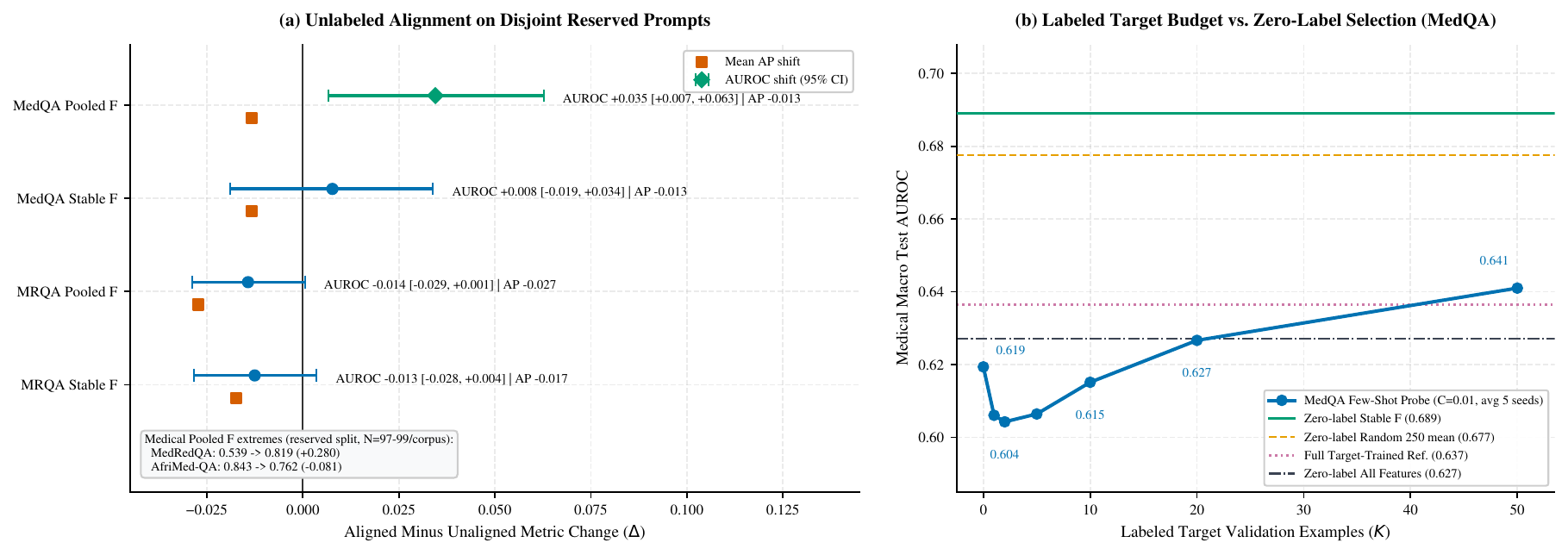}
    \caption{Test-time adaptation using disjoint target access (\textbf{RQ3}). \textbf{(a)} Leakage-free CORAL on untouched reserved adaptation prompts ($N=97$--$99$) raises AUROC for medical Pooled $F$ ($+0.035$, driven by MedRedQA $+0.280$ despite AfriMed-QA $-0.081$) while trading off AP ($-0.013$) and lowering MRQA point estimates. \textbf{(b)} Mixing $K \in \{0..50\}$ labeled medical validation items into pooled training initially hurts at $K \le 10$ and trails zero-label $250$-d selection even at $K=50$.}
    \label{fig:adaptation}
\end{figure*}

\subsection{Disjoint Unlabeled Prompts (Reserved CORAL) and Labeled Target Budgets}

When an unlabeled batch of target prompts (the held-out reserved adaptation split, $97$--$99$ examples medical, $97$ MRQA) is available prior to test evaluation, closed-form CORAL alignment keeps feature selection and $C$ unchanged and yields sharply conditional effects (Figure~\ref{fig:adaptation}(a); per-fold values in Appendix Table~\ref{tab:res_coral}). On medical QA, CORAL lifts \emph{Pooled $F$} AUROC from $0.6565$ to $0.6910$ ($+0.0345$, $[+0.0067, +0.0627]$), driven by a large recovery on MedRedQA ($0.5389 \rightarrow 0.8186$, $+0.2797$, $[+0.1260, +0.4304]$) offset by an $-0.0812$ ($[-0.1242, -0.0415]$) loss on AfriMed-QA and a slight drop in overall mean AP ($0.4399 \rightarrow 0.4266$). When starting from \emph{Stable $F$}, alignment adds an inconclusive $+0.0077$ ($[-0.0189, +0.0339]$, $0.6891 \rightarrow 0.6968$, AP $0.4233 \rightarrow 0.4100$). On MRQA, reserved-batch CORAL lowers point estimates on both \emph{Pooled $F$} ($0.8068 \rightarrow 0.7925$, $[-0.0288, +0.0005]$) and \emph{Stable $F$} ($0.7803 \rightarrow 0.7678$, $[-0.0284, +0.0036]$), consistent with an earlier matched transductive PCA-$250$ check where CORAL dropped AUROC from $0.7778$ to $0.7246$ ($-0.0532$, $[-0.0782, -0.0287]$). Across our two settings, second-order covariance matching rescues off-register consumer QA when applied on top of pooled features but does not consistently improve strong feature selectors.

Similarly, adding small numbers of labeled target validation examples ($K \in \{1..50\}$ over $5$ random draws) to full $2{,}560$-d source training without validation refit (fixed $C=0.01$, target shots upweighted by $N_S/K$, starting at $0.6194$ for $K=0$ vs.\ $0.6272$ with train+val refit) does not beat zero-label coordinate selection on medical QA (Figure~\ref{fig:adaptation}(b)): per-seed average AUROC dips to $0.6043$--$0.6152$ for $K \in \{1..10\}$, reaches $0.6267$ at $K=20$, and climbs to $0.6410$ at $K=50$, still trailing zero-label \emph{Stable $F$} ($0.6891$), \emph{Signed Mean-$t$} ($0.6879$), and even \emph{Random 250} ($0.6775$).

\subsection{Zero-Shot Cross-Task Transfer and Label Sensitivity}

\begin{table}[t]
\centering
\scriptsize
\setlength{\tabcolsep}{3pt}
\resizebox{\columnwidth}{!}{%
\begin{tabular}{lrrrrr}
\toprule
MRQA Error Rule & Errors/650 & All feats. & Pooled $F$ & Mean $F$ & Stable $F$ \\
\midrule
F1 $< 0.25$ & 133 & 0.7324 & 0.7549 & 0.7385 & 0.7477 \\
F1 $< 0.50$ (primary) & 180 & 0.7628 & \textbf{0.8068*} & 0.7986 & 0.7803 \\
F1 $< 0.75$ & 236 & 0.7266 & \textbf{0.7758*} & \textbf{0.7626*} & \textbf{0.7619*} \\
Exact mismatch & 274 & 0.7093 & \textbf{0.7660*} & \textbf{0.7553*} & \textbf{0.7432*} \\
\midrule
\multicolumn{6}{l}{\emph{Zero-Shot Cross-Task Dense Probe (All Features)}} \\
\multicolumn{3}{l}{MedQA (9 corpora) $\rightarrow$ MRQA (5):} & \multicolumn{3}{r}{\textbf{0.4996} \quad [0.4429, 0.5601]} \\
\multicolumn{3}{l}{MRQA (5 corpora) $\rightarrow$ MedQA (9):} & \multicolumn{3}{r}{\textbf{0.4889} \quad [0.4398, 0.5369]} \\
\bottomrule
\end{tabular}%
}
\caption{MRQA test AUROC sensitivity under alternative F1/EM error thresholds (top, freezing primary probes; * marks 95\% CI $>0$ over All features) and zero-shot bidirectional cross-task AUROC (bottom).}
\label{tab:grading_cross}
\end{table}

As shown in Figure~\ref{fig:mrqa_lodo}(c) and Table~\ref{tab:grading_cross}, error probes trained on medical QA average $0.4996$ AUROC on MRQA ($95\%$ CI $[0.4429, 0.5601]$; target range $0.3568$ SearchQA to $0.5794$ HotpotQA), and probes trained on MRQA average $0.4889$ on medical QA ($[0.4398, 0.5369]$; target range $0.3555$ MedRedQA to $0.7387$ iCliniq, an $8$-error target). Applying feature filtering or full CORAL does not rescue zero-shot cross-task transfer ($0.4154$--$0.4849$, Appendix Table~\ref{tab:crosstask}). Conversely, within MRQA, Table~\ref{tab:grading_cross} confirms that \emph{Pooled $F$}'s ranking advantage persists across stricter and looser answer-overlap definitions ($0.7549$--$0.8068$), clearing zero against all features on F1 $< 0.50$, F1 $< 0.75$, and exact mismatch.

\section{Related Work}

\paragraph{Internal Probing and Prediction Timing.}
Post-hoc uncertainty estimators such as answer-conditioned $P(\text{True})$ \citep{kadavath2022mostly} and multi-sample semantic entropy \citep{farquhar2024entropy} score generated outputs after decoding, whereas \citet{azaria2023internal} probe hidden states of statements already supplied or generated in context. Closer to our inference regime, \citet{kadavath2022mostly}'s $P(\text{IK})$ predicts unconditioned question answerability, \citet{ji2024risk} extract pre-response query activations for hallucination risk, and \citet{moreno2025noanswer} train question-only difference-of-means probes predicting self-correctness across knowledge datasets (and noting weaker math generalization). Our work advances pre-generation probing into multi-source medical and extractive LODO protocols, reporting both ranking discrimination (AUROC) and minority-class average precision \citep{davis2006pr} conditional on upcoming greedy model errors.

\paragraph{Transfer Robustness, Sparse Autoencoders, and Controlled Adaptation.}
\citet{mishra2026medical} isolate register, medical specialty, and corpus shifts using candidate-answer verification prompts over rewritten MedQA facts, finding larger drops under corpus shift than specialty or register rewrites. \citet{duan2026transfer} constructs $7\times 7$ correctness transfer matrices across three LLM families at a generated pre-answer marker, showing stronger transfer within task families than across them and introducing a random-direction floor baseline. Similarly, \citet{kantamneni2025sae} show that SAE probes generally underperform $L_2$ logistic regression across shift and scarcity benchmarks. We extend that skepticism using Gemma Scope~2 dictionaries \citep{mcdougall2025scope}, adopt $SD$/$TD$ shift framing from \citet{calderon2024robustness}, and confirm \citet{gulrajani2021domainbed}'s core lesson: rigorous ERM controls ($250$ random coordinates, V-REx $\beta=0$) explain gains often attributed to complex invariance \citep{arjovsky2019irm,krueger2021rex,sagawa2020groupdro} or alignment \citep{sun2016coral} objectives.

\section{Discussion}

Why do feature selection and optimizer regularization carry almost all the benefit in these pre-generation transfer settings? On roughly $1{,}300$ to $2{,}600$ training instances in $2{,}560$ dimensions, moderately penalized linear probes can align with corpus-specific topic and register directions rather than error signals; notably, $14/14$ full $2{,}560$-d LIBLINEAR probes and $7/9$ medical $\beta=0$ probes select the grid floor $C=0.01$, so penalties stricter than $C=0.01$ on all coordinates remain untested. Restricting to $250$ coordinates—even randomly—cuts capacity tenfold ($+0.050$ medical, $+0.026$ MRQA AUROC); gradient-clipped early-stopped Adam ($\beta=0$) has a parallel effect ($+0.063$, $+0.033$). Beyond capacity reduction, MRQA's uniform passage format aligns \emph{Pooled $F$} across sources ($0.905$ mean sign consensus, $+0.043$ AP over Random 250), whereas heterogeneous medical corpora cause unsigned \emph{Stable $F$} to keep sign-reversing features ($10.7\%$ unanimous) while \emph{Signed Mean-$t$} matches top AUROC ($0.688$) and leads AP ($0.441$). Cross-task transfer near $0.50$ reflects joint shifts in passage context, topic, and grading criteria.

\section{Limitations}

This study evaluates one frozen backbone (\texttt{google/gemma-3-4b-it}), layer ($17$), token position, and primary linear family ($L_2$ logistic regression), leaving other models, depths, nonlinear probes, penalties stricter than $C=0.01$, and few-shot adaptation inside selected $250$-d subspaces outside our scope. Medical labels rely on automated \texttt{Qwen2.5-7B-Instruct} comparisons rather than clinicians, and three consumer medical portals contain only $8$ to $10$ test errors, widening per-target intervals. While exact/fuzzy overlap audits ($29$ near-duplicate OpenMedQA/MedQA-open pairs) and paired family holdouts control known lineage, subtle web/exam overlap may persist, and MRQA test rows retain passage correlations across items within split. Intervals condition on fixed corpora and probes, and AUROC/AP measure ranking quality rather than calibrated deferral or rollout-based semantic entropy.

\section{Conclusion}

Across three empirical questions: (1) single-source pre-generation probes degrade under cross-corpus shift in medical QA ($0.624 \rightarrow 0.558$) and MRQA ($0.759 \rightarrow 0.709$); (2) pooling sources and cutting dense dimension to $250$ coordinates raises LODO AUROC and AP over full states, SAE latents, and surface controls, though random-$250$, signed-$t$, and V-REx $\beta=0$ controls show capacity reduction and optimizer regularization—not cross-source variance penalties—drive most gains; and (3) disjoint unlabeled CORAL helps medical pooled selection on AUROC ($+0.035$) while hurting AP and MRQA, few-shot full-space budgets trail zero-label selection, and cross-task transfer averages near chance. Simple matched linear controls thus set a strong bar for test-time probing.

\section*{Ethics and AI-Assistance Disclosure}

All underlying datasets are publicly distributed medical and reading-comprehension research benchmarks (MedQuAD NIH resources, MedicationQA consumer inquiries, de-identified Reddit \texttt{r/AskDocs} threads in MedRedQA and AskD-English, HealthCareMagic/iCliniq consultation logs, USMLE/clinical exam pools in AfriMed-QA, OpenMedQA, and MedQA-open, and MRQA~2019). We report aggregated evaluation statistics without redistributing clinical dialogue text or identifiable metadata. Because linear error ranking achieves $0.689$ macro AUROC in medical QA and sits near chance on multiple patient-forum targets, pre-generation probes must never be deployed as autonomous clinical safety filters.

AI coding and writing assistants were used during research engineering, LaTeX refactoring, and manuscript editing under author supervision. All pipeline scripts, SHA256-checked archives, Kaggle execution runs, statistical bootstrap checks, and tabular metrics were executed and verified directly by the authors.

\section*{Acknowledgments}
We thank the FAST School of Computing at the National University of Computer and Emerging Sciences (FAST-NUCES), Lahore, for institutional support, and acknowledge Kaggle and Google Colab GPU resources used for frozen model extraction and controlled validation runs.

\bibliography{references}

\appendix

\section{Data Accounting, Provenance Audit, and Protocol Matrix}
\label{sec:appendix_a}

\begin{table*}[t]
\centering
\scriptsize
\setlength{\tabcolsep}{4pt}
\resizebox{\textwidth}{!}{%
\begin{tabular}{llrrrrrrr}
\toprule
Corpus A & Corpus B & Exact Matches & Char $\ge 0.95$ & Char $\ge 0.85$ & Char $\ge 0.75$ & Word $\ge 0.80$ & Max Char Cos. & Max Word Cos. \\
\midrule
MedRedQA & AskD-English & 1 & 1 & 1 & 1 & 1 & 1.0000 & 1.0000 \\
OpenMedQA & MedQA-open & 0 & 8 & 29 & 45 & 21 & 0.9840 & 0.9802 \\
MedQuAD & MedicationQA & 0 & 0 & 0 & 0 & 0 & 0.6667 & 0.5559 \\
MedQuAD & AskD-English & 0 & 0 & 0 & 0 & 0 & 0.6455 & 0.2422 \\
MedQuAD & AfriMed-QA & 0 & 0 & 0 & 0 & 0 & 0.6166 & 0.7474 \\
All other 31 pairs & --- & 0 & 0 & 0 & 0 & 0 & $\le 0.6099$ & $\le 0.5492$ \\
\bottomrule
\end{tabular}%
}
\caption{Exact and near-duplicate question similarity across all $36$ Medical QA corpus pairs ($5{,}848$ binary items). The single MedRedQA--AskD-English overlap sits between the MedRedQA reserved adaptation split and AskD-English training split; OpenMedQA and MedQA-open share 29 near-identical USMLE vignettes at char cosine $\ge 0.85$ (prompting our paired family holdout check in Table~\ref{tab:family_holdout}). All 5 MRQA corpora exhibit $0$ cross-split passage or question overlap and $0$ cross-corpus exact overlaps.}
\label{tab:provenance}
\end{table*}

\begin{table*}[t]
\centering
\scriptsize
\setlength{\tabcolsep}{4pt}
\resizebox{\textwidth}{!}{%
\begin{tabular}{p{3.1cm}p{2.0cm}p{3.6cm}p{5.8cm}}
\toprule
Experiment Suite & Benchmarks & Solver / Tuning Rule & Key Protocol Boundary \\
\midrule
Pairwise Transfer & MedQA, MRQA & LR, 3-fold CV on source train & MRQA rerun enforces passage-grouped in-fold scaling ($0.7594$ $TT$ vs.\ $0.6241$ MedQA). \\
Matched LODO & MedQA, MRQA & LR/SVM/Ridge, val macro AUC & Identical $C \in \{0.01..10\}$, train+val refit, convergence up to $20{,}000$ iter. \\
Reserved-Batch CORAL & MedQA, MRQA & Tuned $C$ + selected coords & $\gamma=1.0$ fitted strictly on untouched reserved adaptation prompts ($N=97$--$99$). \\
V-REx \& $\beta=0$ Controls & MedQA, MRQA & Adam ($\text{lr}=10^{-3}$), grad clip $1.0$ & Tuned $\beta \ge 0$ vs.\ $\beta=0$ ($150$ ep., tuned $C$ MedQA; $100$ ep., $\text{wd}=10^{-4}$ MRQA). \\
Paired Family Holdouts & MedQA (4 targ.) & Same as Matched LODO & Drops both AskDocs (MedRedQA+AskD) or both MedQA (OpenMedQA+MedQA-open). \\
Exploratory Suite & MedQA (App.~\ref{sec:appendix_c}) & Fixed $C$ or custom losses & Transductive test-cov CORAL ($C=1$), Few-Shot val ($C=0.01$), DEAP, MoP, LRCP, DPO. \\
\bottomrule
\end{tabular}%
}
\caption{Audit separating matched protocols in the main text from exploratory adaptation runs in Appendix~\ref{sec:appendix_c}.}
\label{tab:protocols}
\end{table*}

Across all $5{,}850$ raw medical items ($650$ per corpus), schema inspection confirms zero stored output-token logits or log-probabilities in either benchmark's saved parquet metadata ($14$ medical metadata columns; $13$ MRQA record columns). Every row vector in the audited medical LODO archive matches its raw \texttt{.npy} counterpart bit-for-bit by SHA256. Table~\ref{tab:provenance} summarizes pairwise similarity, and Table~\ref{tab:protocols} documents protocol distinctions across every table.

\section{Complete Per-Corpus Results, Sign Diagnostics, and Cross-Task Transfer}
\label{sec:appendix_b}

Tables~\ref{tab:per_fold_auc}, \ref{tab:medqa_ap}, \ref{tab:family_holdout}, \ref{tab:sign_diag}, and \ref{tab:crosstask} report full per-corpus test AUROCs and Average Precision values across all 9 medical and 5 MRQA LODO folds, paired provenance holdouts, source-wise sign-agreement diagnostics, and target-level cross-task transfer AUROCs.

\begin{table*}[t]
\centering
\scriptsize
\setlength{\tabcolsep}{3.2pt}
\resizebox{\textwidth}{!}{%
\begin{tabular}{lrrrrrrrrr}
\toprule
Target Corpus & Err./N & All feats. & Random 250 & PCA 250 & Pooled $F$ & Mean $F$ & Stable $F$ & Signed Mean-$t$ & TF-IDF (fixed) \\
\midrule
MedQuAD & 42/131 & 0.6808 & 0.7691 & 0.7175 & 0.7592 & \textbf{0.7889} & 0.6752 & 0.7804 & 0.4961 \\
MedicationQA & 30/129 & 0.6074 & 0.7310 & 0.6620 & 0.7684 & 0.7589 & \textbf{0.8205} & 0.7902 & 0.6146 \\
MedRedQA & 8/130 & 0.4355 & \textbf{0.6510} & 0.5389 & 0.5389 & 0.5799 & 0.5686 & 0.5727 & 0.4457 \\
AfriMed-QA & 35/130 & 0.7702 & 0.8018 & 0.7777 & \textbf{0.8430} & 0.7934 & 0.7979 & 0.8099 & 0.4794 \\
OpenMedQA & 93/130 & 0.7042 & 0.6922 & 0.6821 & 0.6786 & \textbf{0.7140} & 0.6643 & 0.7030 & 0.6414 \\
MedQA-open & 74/130 & 0.5931 & 0.6118 & 0.6127 & 0.6310 & 0.6349 & \textbf{0.6658} & 0.6404 & 0.6496 \\
HealthCareMagic & 19/130 & 0.4438 & 0.5009 & 0.4647 & 0.4941 & 0.5069 & 0.5747 & 0.5330 & \textbf{0.5827} \\
AskD-English & 10/130 & 0.5767 & 0.5467 & 0.5575 & 0.4958 & 0.5400 & 0.6150 & 0.5908 & \textbf{0.6867} \\
iCliniq & 8/130 & \textbf{0.8330} & 0.7927 & 0.7582 & 0.6998 & 0.7234 & 0.8197 & 0.7705 & 0.6547 \\
\midrule
MedQA Mean (9) & 319/1,170 & 0.6272 & 0.6775 & 0.6413 & 0.6565 & 0.6711 & \textbf{0.6891} & 0.6879 & 0.5834 \\
\midrule
HotpotQA & 24/130 & 0.6474 & 0.6805 & 0.6262 & \textbf{0.7571} & 0.7323 & 0.6769 & 0.7543 & 0.4894 \\
NaturalQuestions & 46/130 & 0.8093 & 0.8049 & 0.8131 & \textbf{0.8577} & 0.8450 & 0.7943 & 0.8442 & 0.6017 \\
SQuAD & 11/130 & 0.7586 & \textbf{0.8179} & 0.7998 & 0.8144 & 0.8174 & 0.7991 & 0.8037 & 0.5875 \\
SearchQA & 72/130 & 0.7548 & 0.7938 & \textbf{0.8103} & 0.7900 & 0.7792 & 0.8036 & 0.7579 & 0.5170 \\
TriviaQA & 27/130 & 0.8439 & \textbf{0.8447} & 0.8393 & 0.8148 & 0.8191 & 0.8278 & 0.8288 & 0.5189 \\
\midrule
MRQA Mean (5) & 180/650 & 0.7628 & 0.7883 & 0.7778 & \textbf{0.8068} & 0.7986 & 0.7803 & 0.7978 & 0.5429 \\
\bottomrule
\end{tabular}%
}
\caption{Complete per-corpus test AUROCs across all 9 Medical QA and 5 MRQA LODO folds under source-only validation tuning.}
\label{tab:per_fold_auc}
\end{table*}

\begin{table*}[t]
\centering
\scriptsize
\setlength{\tabcolsep}{3.2pt}
\resizebox{\textwidth}{!}{%
\begin{tabular}{lrrrrrrrrr}
\toprule
Target Corpus & Err.\ Rate & All feats.\ AP & Random 250 AP & Pooled $F$ AP & Mean $F$ AP & Stable $F$ AP & Signed Mean-$t$ AP & Pool SAE AP & Targ.\ Dense AP \\
\midrule
MedQuAD & 0.3206 & 0.4303 & 0.5496 & 0.5468 & \textbf{0.6062} & 0.4385 & 0.5766 & 0.3977 & 0.5910 \\
MedicationQA & 0.2326 & 0.3478 & 0.4725 & 0.4993 & 0.4694 & \textbf{0.5702} & 0.5401 & 0.3379 & 0.4710 \\
MedRedQA & 0.0615 & 0.0599 & 0.1055 & 0.1386 & 0.0853 & 0.0735 & 0.0869 & \textbf{0.2419} & 0.1307 \\
AfriMed-QA & 0.2692 & 0.5431 & 0.5972 & \textbf{0.6982} & 0.5476 & 0.5944 & 0.5994 & 0.5720 & 0.6378 \\
OpenMedQA & 0.7154 & 0.7900 & 0.8083 & 0.8069 & 0.8234 & 0.7925 & \textbf{0.8286} & 0.7646 & 0.7547 \\
MedQA-open & 0.5692 & 0.6374 & 0.6528 & 0.7106 & 0.6838 & 0.7047 & 0.7066 & \textbf{0.7195} & 0.6853 \\
HealthCareMagic & 0.1462 & 0.1397 & 0.1816 & 0.2366 & 0.2147 & 0.2078 & \textbf{0.2528} & 0.1921 & 0.1726 \\
AskD-English & 0.0769 & 0.1194 & 0.1216 & 0.0976 & 0.1160 & 0.1252 & 0.1276 & 0.0939 & \textbf{0.1364} \\
iCliniq & 0.0615 & 0.3068 & 0.2985 & 0.2246 & 0.2244 & 0.3032 & 0.2458 & \textbf{0.4093} & 0.1325 \\
\midrule
MedQA Mean (9) & 0.2725 & 0.3749 & 0.4208 & 0.4399 & 0.4190 & 0.4233 & \textbf{0.4405} & 0.4143 & 0.4124 \\
\bottomrule
\end{tabular}%
}
\caption{Per-corpus Average Precision (AP) on the 9 medical LODO folds. On MRQA, macro APs are $0.5521$ (All features), $0.5987$ (Random 250), $\mathbf{0.6420}$ (Pooled $F$), $0.6179$ (Mean $F$), $0.5749$ (Stable $F$), $0.6229$ (Signed Mean-$t$), and $0.4914$ (Pooled SAE).}
\label{tab:medqa_ap}
\end{table*}

\begin{table}[t]
\centering
\scriptsize
\setlength{\tabcolsep}{3pt}
\resizebox{\columnwidth}{!}{%
\begin{tabular}{lrrrrrrrr}
\toprule
Target & Full 1 & Full 2 & Pool-$F$ 1 & Pool-$F$ 2 & Mean-$F$ 1 & Mean-$F$ 2 & Stb-$F$ 1 & Stb-$F$ 2 \\
\midrule
MedRedQA & .4355 & .5215 & .5389 & .5656 & .5799 & .6311 & .5686 & .6250 \\
AskD-En. & .5767 & .5758 & .4958 & .4908 & .5400 & .5750 & .6150 & .5300 \\
OpenMedQA & .7042 & .6585 & .6786 & .6829 & .7140 & .6542 & .6643 & .6332 \\
MedQA-open & .5931 & .5907 & .6310 & .6112 & .6349 & .6388 & .6658 & .6532 \\
\midrule
Mean (4) & .5774 & .5867 & .5861 & .5876 & .6172 & .6248 & .6284 & .6104 \\
\bottomrule
\end{tabular}%
}
\caption{Paired family holdout AUROC when holding out one (1) vs.\ both (2) related corpora sharing AskDocs (MedRedQA+AskD-English) or MedQA (OpenMedQA+MedQA-open) lineage.}
\label{tab:family_holdout}
\end{table}

\begin{table}[t]
\centering
\scriptsize
\setlength{\tabcolsep}{3pt}
\resizebox{\columnwidth}{!}{%
\begin{tabular}{llcccc}
\toprule
Bench. & Selector ($250$) & $|\overline{\operatorname{sign}}(t_e)|$ & Unan.\ Sign & Overlap Stb-$F$ & Overlap Pool-$F$ \\
\midrule
MedQA & Pooled $F$ & 0.5993 & 23.1\% & 21.0 & 250.0 \\
MedQA & Stable $F$ (unsign.) & \textbf{0.4651} & \textbf{10.7\%} & 250.0 & 21.0 \\
MedQA & Signed Mean-$t$ & 0.8112 & 42.7\% & 24.8 & 76.9 \\
MedQA & Signed Stable-$F$ & 0.8871 & 58.0\% & 54.1 & 57.8 \\
\midrule
MRQA & Pooled $F$ & 0.9048 & 82.6\% & 40.4 & 250.0 \\
MRQA & Stable $F$ (unsign.) & 0.6740 & 50.2\% & 250.0 & 40.4 \\
MRQA & Signed Mean-$t$ & 0.9716 & 94.3\% & 46.0 & 144.0 \\
MRQA & Signed Stable-$F$ & 1.0000 & 100.0\% & 120.4 & 78.2 \\
\bottomrule
\end{tabular}%
}
\caption{Mean cross-source Welch $t$ sign agreement across selected $250$-feature subsets. Unsigned Stable $F$ drops to $10.7\%$ unanimous direction on MedQA.}
\label{tab:sign_diag}
\end{table}

\begin{table}[ht]
\centering
\scriptsize
\setlength{\tabcolsep}{3.2pt}
\resizebox{\columnwidth}{!}{%
\begin{tabular}{lccc}
\toprule
Direction / Target & All features & CORAL Full & Stable-$F$ \\
\midrule
\multicolumn{4}{l}{\emph{MedQA (9 corpora) $\rightarrow$ MRQA (Pooled-$F$ mean: 0.4849; CORAL ANOVA: 0.4426)}} \\
SQuAD & 0.5225 & 0.4102 & 0.4163 \\
TriviaQA & 0.5548 & 0.5135 & 0.5743 \\
SearchQA & 0.3568 & 0.4162 & 0.3391 \\
HotpotQA & 0.5794 & 0.4029 & 0.3518 \\
NaturalQuestions & 0.4842 & 0.5567 & 0.3957 \\
Mean (MedQA $\rightarrow$ MRQA) & 0.4996 & 0.4599 & 0.4154 \\
\midrule
\multicolumn{4}{l}{\emph{MRQA (5 corpora) $\rightarrow$ MedQA}} \\
MedQuAD & 0.5342 & 0.4743 & 0.3135 \\
MedicationQA & 0.5481 & 0.5192 & 0.4327 \\
MedRedQA & 0.3555 & 0.4119 & 0.5492 \\
AfriMed-QA & 0.4529 & 0.4632 & 0.1832 \\
OpenMedQA & 0.4380 & 0.5161 & 0.3964 \\
MedQA-open & 0.4011 & 0.3974 & 0.3972 \\
HealthCareMagic & 0.4087 & 0.3509 & 0.5737 \\
AskD-English & 0.5225 & 0.5525 & 0.5842 \\
iCliniq & 0.7387 & 0.5994 & 0.6947 \\
Mean (MRQA $\rightarrow$ MedQA) & 0.4889 & 0.4761 & 0.4583 \\
\bottomrule
\end{tabular}%
}
\caption{Per-target zero-shot cross-task transfer AUROCs.}
\label{tab:crosstask}
\end{table}

\section{Reserved-Batch CORAL and Exploratory Adaptation Archive}
\label{sec:appendix_c}

Table~\ref{tab:res_coral} reports per-corpus AUROC and AP shifts for disjoint reserved-batch CORAL across both benchmarks. Table~\ref{tab:exploratory} records all earlier exploratory medical variants executed under distinct regularization or objective formulations: (i) transductive test-covariance CORAL at fixed $C=1.0$ ($0.6130$ full, $0.6420$ PCA-250, $0.6541$ pooled ANOVA-250 vs.\ unweighted base $0.5755$), (ii) Calibrated V-REx per-fold $\beta^*$ vs.\ exact $\beta=0$ control, (iii) Few-Shot validation adaptation ($K=50$ reaching $0.6410$ vs.\ $K=0$ $0.6194$ at $C=0.01$; MRQA ensembled logits across 5 validation draws yield $0.7715$, $0.7801$, $0.7825$ for $K=10, 20, 50$), (iv) DEAP epistemic distance probability calibration (mean Brier score improves from $0.2384$ to $0.2042$, $\Delta = +0.0342$; AUROC $0.5755 \rightarrow 0.6180$), (v) Mixture-of-Probes (softmax routing over 8 source probes at $T=1.0$, $0.6472$ AUROC, $+0.2387$ MedRedQA vs.\ $-0.2059$ iCliniq), (vi) Risk-constrained LRCP ($0.5953$ AUROC vs.\ own $0.5953$ base; false high-error-confidence rate on correct answers, $\hat{p} > 0.75 \mid y=0$, reduced from $0.1940$ to $0.1773$), (vii) DPO pairwise ranking probe ($0.5955$ AUROC), and (viii) RBF Kernel $\text{MMD}^2$ distance vs.\ Target-Drop correlation ($r=0.2509$, $p=0.515$; Spearman $\rho=0.3500$, $p=0.356$). Earlier exploratory medical ANOVA validation sweeps across $K \in \{50,100,250\}$ scored $0.6790$ (dense) and $0.6599$ (SAE), which motivated fixing $K=250$ across all matched evaluations.

\begin{table*}[t]
\centering
\scriptsize
\setlength{\tabcolsep}{4pt}
\resizebox{\textwidth}{!}{%
\begin{tabular}{lrrrrrrr}
\toprule
Corpus ($N_{\text{res}}$) & Pooled $F$ Orig & Pooled $F$ CORAL & Pooled $F$ AP ($\Delta$AP) & Stable $F$ Orig & Stable $F$ CORAL & Stable $F$ $\Delta$AUC & Pooled $F$ $\Delta$AUC [95\% CI] \\
\midrule
MedQuAD (99) & .7592 & .7408 & .5524 ($+.006$) & .6752 & .7004 & $+.0251$ & $-.0185$ $[-.0725, +.0329]$ \\
MedicationQA (97) & .7684 & .7360 & .4455 ($-.054$) & .8205 & .7721 & $-.0485$ & $-.0323$ $[-.0852, +.0189]$ \\
MedRedQA (97) & .5389 & .8186 & .1771 ($+.039$) & .5686 & .7674 & $+.1988$ & $\mathbf{+.2797\ [+.1260,\ +.4304]}$ \\
AfriMed-QA (97) & .8430 & .7618 & .4979 ($-.200$) & .7979 & .7242 & $-.0737$ & $\mathbf{-.0812\ [-.1242,\ -.0415]}$ \\
OpenMedQA (97) & .6786 & .6986 & .8471 ($+.040$) & .6643 & .6617 & $-.0026$ & $+.0201$ $[-.0366, +.0776]$ \\
MedQA-open (97) & .6310 & .6412 & .6931 ($-.018$) & .6658 & .6655 & $-.0002$ & $+.0101$ $[-.0280, +.0521]$ \\
HealthCareMagic (97) & .4941 & .5595 & .2542 ($+.018$) & .5747 & .5889 & $+.0142$ & $+.0654$ $[-.0294, +.1731]$ \\
AskD-English (97) & .4958 & .5425 & .0968 ($-.001$) & .6150 & .5517 & $-.0633$ & $+.0467$ $[-.0600, +.1592]$ \\
iCliniq (97) & .6998 & .7203 & .2751 ($+.050$) & .8197 & .8391 & $+.0195$ & $+.0205$ $[-.0738, +.1158]$ \\
\midrule
MedQA Mean (9) & .6565 & .6910 & .4266 ($-.013$) & .6891 & .6968 & $+.0077$ & $\mathbf{+.0345\ [+.0067,\ +.0627]}$ \\
\midrule
SQuAD (97) & .8144 & .8090 & .3904 ($-.040$) & .7991 & .7723 & $-.0267$ & $-.0053$ $[-.0382, +.0298]$ \\
TriviaQA (97) & .8148 & .8022 & .5380 ($-.090$) & .8278 & .7994 & $-.0284$ & $-.0126$ $[-.0478, +.0255]$ \\
SearchQA (97) & .7900 & .7874 & .8321 ($-.011$) & .8036 & .8175 & $+.0139$ & $-.0026$ $[-.0261, +.0199]$ \\
HotpotQA (97) & .7571 & .7303 & .5554 ($+.022$) & .6769 & .6914 & $+.0145$ & $-.0267$ $[-.0696, +.0094]$ \\
Nat.\ Questions (97) & .8577 & .8336 & .7582 ($-.017$) & .7943 & .7583 & $-.0360$ & $-.0241$ $[-.0518, +.0000]$ \\
\midrule
MRQA Mean (5) & .8068 & .7925 & .6148 ($-.027$) & .7803 & .7678 & $-.0125$ & $-.0143$ $[-.0288, +.0005]$ \\
\bottomrule
\end{tabular}%
}
\caption{Per-corpus AUROC and AP for closed-form CORAL fit strictly on disjoint reserved adaptation prompts ($\gamma=1.0$, identical $250$-feature coordinates and tuned $C$). Bold entries mark 95\% CIs excluding zero.}
\label{tab:res_coral}
\end{table*}

\begin{table*}[t]
\centering
\scriptsize
\setlength{\tabcolsep}{3.2pt}
\resizebox{\textwidth}{!}{%
\begin{tabular}{lrrrrrrrrr}
\toprule
Target Corpus & V-REx ($\beta^*$) & V-REx ($\beta=0$) & Tr.\ Base ($C=1$) & Tr.\ Full & Tr.\ PCA & Tr.\ ANOVA & FS $K=50$ & DEAP (Raw Brier $\rightarrow$ DEAP) & MoP \\
\midrule
MedQuAD & 0.7405 (10) & 0.7549 & 0.6043 & 0.5984 & 0.6322 & 0.6501 & 0.6912 & 0.5824 (.348 $\rightarrow$ .306) & 0.5222 \\
MedicationQA & 0.7842 (1) & 0.7990 & 0.5269 & 0.6360 & 0.6650 & 0.6848 & 0.7034 & 0.5128 (.345 $\rightarrow$ .258) & 0.7269 \\
MedRedQA & 0.6834 (0) & 0.6834 & 0.4129 & 0.5256 & 0.7018 & 0.7275 & 0.4775 & 0.5881 (.138 $\rightarrow$ .123) & 0.6742 \\
AfriMed-QA & 0.7883 (10) & 0.7949 & 0.7203 & 0.7080 & 0.6830 & 0.7176 & 0.8099 & 0.6941 (.292 $\rightarrow$ .227) & 0.7627 \\
OpenMedQA & 0.7219 (10) & 0.7047 & 0.6911 & 0.6716 & 0.6222 & 0.6472 & 0.7073 & 0.7152 (.261 $\rightarrow$ .244) & 0.7268 \\
MedQA-open & 0.6412 (0) & 0.6412 & 0.6057 & 0.6047 & 0.6103 & 0.6166 & 0.5963 & 0.6759 (.313 $\rightarrow$ .262) & 0.6561 \\
HealthCareMagic & 0.5012 (10) & 0.5315 & 0.4282 & 0.4742 & 0.5078 & 0.5192 & 0.5037 & 0.5168 (.248 $\rightarrow$ .233) & 0.4889 \\
AskD-English & 0.4800 (10) & 0.6267 & 0.5033 & 0.5950 & 0.6458 & 0.6250 & 0.5785 & 0.5408 (.141 $\rightarrow$ .124) & 0.6400 \\
iCliniq & 0.6783 (0) & 0.6783 & 0.6865 & 0.7039 & 0.7100 & 0.6988 & 0.7014 & 0.7357 (.059 $\rightarrow$ .062) & 0.6270 \\
\midrule
Macro Mean & 0.6688 & \textbf{0.6905} & 0.5755 & 0.6130 & 0.6420 & 0.6541 & 0.6410 & 0.6180 (.2384 $\rightarrow$ .2042) & 0.6472 \\
\bottomrule
\end{tabular}%
}
\caption{Matched 9-fold V-REx ($\beta^*$ vs.\ $\beta=0$) alongside earlier exploratory medical adaptation runs (transductive CORAL using pooled ANOVA~$F$, Few-Shot $K=50$, DEAP AUROC and Brier calibration, Mixture-of-Probes). Additional exploratory means: LRCP $0.5953$ AUROC (false high-error-confidence rate $\hat{p}>0.75\mid y=0$ reduced by $+0.0166$), DPO ranking $0.5955$, Kernel $\text{MMD}^2$ vs.\ Target-Drop $r=0.2509$ ($p=0.515$).}
\label{tab:exploratory}
\end{table*}

\section{Prompt Templates, Extraction Details, and Judge Rubric}
\label{sec:appendix_d}

\paragraph{Gemma-3 Extraction and Decoding.}
All activations and responses use \texttt{google/gemma-3-4b-it} (revision \texttt{093f9f38\allowbreak 8b31de27\allowbreak 6ce2de16\allowbreak 4bdc2081\allowbreak 324b9767}) in BF16 eager attention with deterministic greedy decoding (\texttt{do\_sample=False}) and \texttt{max\_prompt\_\allowbreak tokens=3072} (head $75\%$ + tail $25\%$ token trim when exceeded). Medical prompts format user content with: \textit{``Answer the medical question directly and accurately. Give only the medically relevant answer. Be concise and use at most about 100 words. Do not discuss the dataset or evaluation process. If the information is insufficient, say so. \textbackslash n\textbackslash nQuestion: \{question\}''} with generation cap $256$ tokens across all nine medical corpora. MRQA prompts use: \textit{``Answer the question concisely using only the provided context. Give a short direct answer span.\textbackslash n\textbackslash nContext: \{context\}\textbackslash n\textbackslash nQuestion: \{question\}\textbackslash n\textbackslash nAnswer:''} with generation cap $48$ tokens. Gemma Scope~2 activations are extracted via \texttt{gemma-scope-2-4b-it-res} / \texttt{layer\_17\_width\_\allowbreak 16k\_l0\_medium} at the exact prefill sequence length immediately prior to generation.

\paragraph{Automated Medical Judge Rubric (Qwen2.5-7B-Instruct).}
Each medical completion is scored deterministically (\texttt{do\_sample=False}, \texttt{max\_new\_tokens=100} for seven corpora and $128$ for AfriMed-QA and OpenMedQA; JSON output {\footnotesize\texttt{\{"label": "correct|incorrect|uncertain", "reason": "..."\}}}) against the corpus reference(s) under the fixed instruction to label \texttt{correct} when materially consistent with core medical facts without penalizing concise paraphrases, \texttt{incorrect} when materially wrong, contradicting essential facts, or omitting core information, and \texttt{uncertain} when references are insufficient or ambiguous. Unparseable outputs default conservatively to \texttt{uncertain} and are excluded from binary labels.

\end{document}